\documentclass[letterpaper, 10 pt, conference]{ieeeconf}  % Comment this line out if you need a4paper

\IEEEoverridecommandlockouts                              % This command is only needed if 
\usepackage{cite}
\usepackage{amsmath,amssymb,amsfonts}
\usepackage{graphicx}
\usepackage[table]{xcolor} % for the table
\usepackage{multirow} % for the table
\usepackage{makecell} % for the table
\usepackage{float}
\usepackage{pax}
\usepackage{hyperref}
\usepackage{url} 
\usepackage{lipsum}
\usepackage{siunitx}
\usepackage{subcaption}

\usepackage{color,soul}

\title{\LARGE \bf
Self-Supervised Monocular Visual Drone Model Identification through Improved Occlusion Handling
}

\author{Stavrow A. Bahnam$^{1}$, Christophe De Wagter$^{1}$,  Guido C.H.E. de Croon$^{1}$%, <-this % stops a space
\thanks{*This work was not supported by any organization}% <-this % stops a space
\thanks{$^{1}$The authors are with the Micro Air Vehicle Lab of the Faculty
of Aerospace Engineering, Delft University of Technology, 2629 HS
Delft, The Netherlands {\tt\small S.A.Bahnam@tudelft.nl, 
C.deWagter@tudelft.nl,
G.C.H.E.deCroon@tudelft.nl}}%
}

\begin{document}

\maketitle
\thispagestyle{empty}
\pagestyle{empty}

%%%%%%%%%%%%%%%%%%%%%%%%%%%%%%%%%%%%%%%%%%%%%%%%%%%%%%%%%%%%%%%%%%%%%%%%%%%%%%%%
\begin{abstract}

Ego-motion estimation is vital for drones when flying in GPS-denied environments. Vision-based methods struggle when flight speed increases and close-by objects lead to difficult visual conditions with considerable motion blur and large occlusions. To tackle this, vision is typically complemented by state estimation filters that combine a drone model with inertial measurements. However, these drone models are currently learned in a supervised manner with ground-truth data from external motion capture systems, limiting scalability to different environments and drones. In this work, we propose a self-supervised learning scheme to train a neural-network-based drone model using only onboard monocular video and flight controller data (IMU and motor feedback). We achieve this by first training a self-supervised relative pose estimation model, which then serves as a teacher for the drone model. To allow this to work at high speed close to obstacles, we propose an improved occlusion handling method for training self-supervised pose estimation models. Due to this method, the root mean squared error of resulting odometry estimates is reduced by an average of $15\%$. Moreover, the student neural drone model can be successfully obtained from the onboard data. It even becomes more accurate at higher speeds compared to its teacher, the self-supervised vision-based model. We demonstrate the value of the neural drone model by integrating it into a traditional filter-based VIO system (ROVIO), resulting in superior odometry accuracy on aggressive 3D racing trajectories near obstacles. Self-supervised learning of ego-motion estimation represents a significant step toward bridging the gap between flying in controlled, expensive lab environments and real-world drone applications. The fusion of vision and drone models will enable higher-speed flight and improve state estimation, on any drone in any environment.

\end{abstract}

%%%%%%%%%%%%%%%%%%%%%%%%%%%%%%%%%%%%%%%%%%%%%%%%%%%%%%%%%%%%%%%%%%%%%%%%%%%%%%%%

\section{introduction}

Accurate ego-motion estimation is crucial for controlling autonomous drones. However, this is still challenging in GPS-denied areas. An often employed approach is to have the drone perform Visual Inertial Odometry (VIO), which makes use of a camera and an inertial measurement unit (IMU) to determine its velocity and rotation rates \cite{vio_general}. Since VIO only requires lightweight sensors (a camera and IMU), it is suitable for drones that have Size Weight and Power (SWaP) restrictions, like Micro Air Vehicles (MAVs).

% TODO: change figure.
\begin{figure}[htb]
    \centering
    \includegraphics[width=0.99\linewidth]{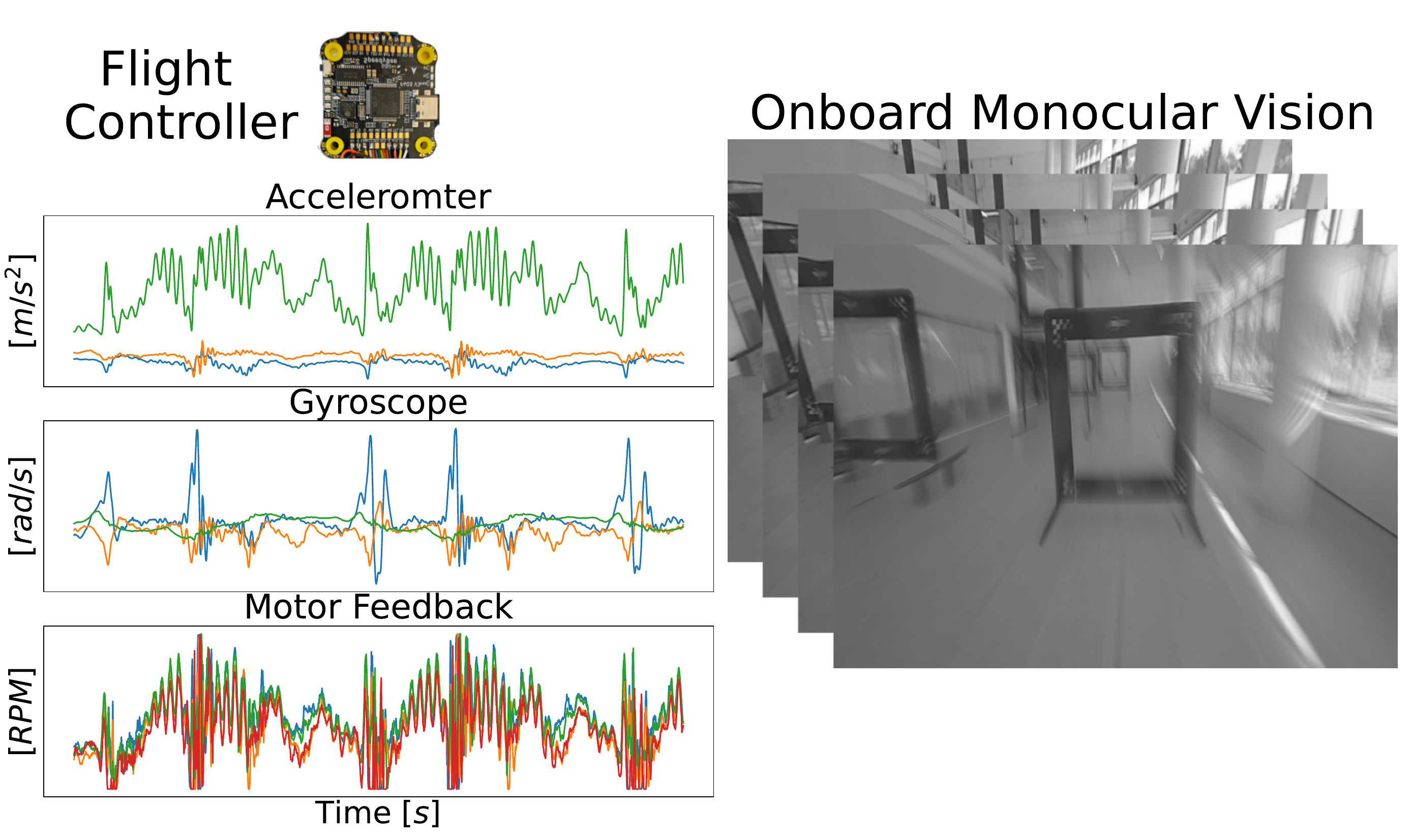}
    \caption{We introduce a self-supervised learning approach that only uses data from onboard the drone to learn ego-motion estimation. Specifically, this data consists of the inertial measurements and control commands available in the flight controller (left), and monocular camera images (right). The approach does not require any external infrastructure and allows to scale the monocular vision-based estimates. 
    %The odometry accuracy from the self-learned odometry is competitive with the state-of-the-art visual-inertial odometry method ROVIO (red line). 
    % We modify the self-supervised pipeline to enable effective vision-based learning on agile flight trajectories, where fast motion near objects leads to motion blur and large (dis-)occlusions. The vision network acts as a teacher of a student network that learns a drone model. This student drone model outperforms its teacher at faster flight velocities.}
    }
    %\caption{Self-supervised drone model from monocular video that estimates one racing lap with an RMSE of $2.0\ m$, while only using flight controller data and $135$ neurons.}
    \label{fig:fig1}
\end{figure}

For certain drone applications, the drone requires a high flight speed. For example, for search and rescue, it would be desirable to find the victims as fast as possible and to cover the maximum amount of area with a single battery. Fast flight does make visual navigation more difficult, as it leads directly to a higher optical flow and motion blur in images. The effects of occlusions also increase when flying very fast near obstacles, as this increases the portion of pixels that cannot be matched from one frame to the next.  
This is problematic for traditional VIO methods that detect and track features across frames like ROVIO \cite{ROVIO}. Learning-based VIO has the potential to be more robust to fast-flight conditions, leveraging the capability of deep neural networks to cope with degraded images, cf. \cite{xu2021cnn,dif_VIO_wagstaff,xu2025cuahn}. However, learning-based VIOs are not as accurate as traditional VIO methods yet and are computationally more expensive, requiring an onboard computer with a GPU, like an NVidia Jetson Orin NX.

Autonomous drone racing is the perfect scenario to test the performance of state estimation algorithms during fast motion. While detection of racing gates can be used to infer the drone's absolute position on the track, gates may not be visible during substantial parts of the flight. Hence, odometry based on ego-motion estimation is important for continuous position feedback control. In the AlphaPilot 2019 race, the runner-up solution employed a filter-based visual-inertial odometry method based on ROVIO \cite{ROVIO}. However, the authors noted in \cite{race_rovio_breaks}  that at higher speeds VIO tracking failures resulted in crashes. The winning solution used \emph{model}-based inertial odometry, combining a drone model with IMU measurements \cite{mavlab-winning}. 
This type of model is not as accurate as VIO, but was shown to increase the long-term stability of the motion-prediction.
%at lower speeds but becomes increasingly useful at higher speeds.
%CDW: This is the finding of this paper: previous paper only showed better long-term prediction, not a relation with speed.
Such a drone model is typically estimated based on data with available ground-truth positions and velocities, as obtained using a motion capture system. For example, based on motion tracking data, in \cite{learned_io} a quadrotor model was learned and fused with an inertial odometry model to get accurate state estimation. Besides the need for an external motion tracking system, the use of global coordinates also limits the approach in \cite{learned_io} to known trajectories.

This limitation also applies to Swift \cite{champion-race}, the seminal autonomous drone system that for the first time beat human FPV (First Person View) drone racing champions. The state estimation of Swift consists of two parts: VIO from an Intel RealSense T265 \cite{realsense} and gate detection to correct for the drift. Furthermore, the authors model a Gaussian noise of the VIO, by fitting it through the position residual obtained from external motion capture measurements to refine the training of their control policy. Visual odometry can behave differently in different environments, due to changes in illumination and the amount of texture. Therefore, for a new environment the control of Swift would need to be retrained with a newly estimated Gaussian noise to keep up the same performance. However, motion capture is not available everywhere.
In another recent study, reinforcement learning was used to directly map racing-gate-segmentation images to control commands for drone racing  \cite{geles2024demonstrating}. The trained neural network successfully exploited the specific properties of the segmented racing gates to seamlessly transfer from simulation to an identical track in the real world. The generalization of this method to other applications will be limited, though.

%This does not lead to VIO that can be applied in multiple situations.

%The downside of this method is that it overfits a trajectory. Furthermore, it requires ground truth position data from a motion capture system, which limits its practical applicability. % GdC: Weet niet goed wat je hiermee wilt hier - learned_io. Is het ook een model-based odometry? 

%Previous approaches used model-based inertial odometry and filter-based visual-inertial odometry, combined with gate detection, for state estimation for autonomous drone racing. Model-based inertial odometry is more robust than filter-based visual-inertial odometry because losing tracked features often results in a diverging filter for VIO. However, model-based inertial odometry usually has a higher drift in position and requires detecting gates frequently to correct for that. As we have seen at AlphaPilot 2019, robust state estimation is more important. The winning team used a robust model-based drag model for quadrotors \cite{mavlab-winning}. While the second-placed team used a more accurate state estimation called ROVIO \cite{ROVIO}. However, the overall filter solution they used turned out to be the bottleneck at high speeds and resulted in a crash \cite{race_rovio_breaks}.
%CDW: I think ROVIO worked but their secondary Kalman filter failed: in any case, be more vague about why unless you cite their exact words from a paper.

In contrast, humans can learn to estimate the state of a drone from only monocular FPV video stream combined with own control inputs. After getting used to a particular drone, this experience seamlessly transfers to new environments, including a-priori unknown environments.
Motivated by this, in this article, we train a self-supervised visual ego estimation network for autonomous drone racing using a monocular video only, such that it can learn in any environment. To also make this possible in racing environments at very high speeds close to obstacles, we first improve existing self-supervised learning (SSL) methods for 'pose' estimation (the translation and rotation between two images). In particular, we propose a novel approach for dealing with parts of the environment that are occluded and with pixels moving out of view - phenomena that are exacerbated by agile drone flight. Moreover, 
%in order to go beyond the perception of human FPV pilots, 
we investigate the self-supervised learning of a neural drone model that uses inertial measurements and motor feedback as inputs. We show that the learned drone model can predict accurate velocities and can complement the visual ego-motion estimates in visually challenging conditions. %Finally, we show that the learned quadrotor model forms a reliable source for learning the scaling parameter of PoseNet.

Our main contributions are:

\begin{itemize}
    %\item Show two cases where the loss functions of state-of-the-art self-supervised visual odometry networks contain errors.
    \item We propose a modified combination of the photometric and depth loss functions in the self-supervised learning of ego-motion to handle occluded areas better
    \item We train for the first time a self-supervised drone model based on images and onboard sensor data, that outperforms its teacher (PoseNet) in velocity estimation. % @ stavrow to be shown!
    % \item \todo{We show that the self-supervised learned drone model in turn forms a reliable source for learning scale consistency in PoseNet (teacher)}
    \item We show that the learned drone model can be successfully integrated into an odometry pipeline, leading to accurate high-speed racing trajectory estimation.
\end{itemize}

Section \ref{section:method} presents the PoseNet, the learned quadrotor model and describes the occlusion problem and the proposed solution.
Section \ref{section:results} presents the performance of the various networks.
Finally, the implications and future work are discussed in Section \ref{section:discussion}.

\begin{figure*}
    \centering
    \includegraphics[width=0.72\linewidth]{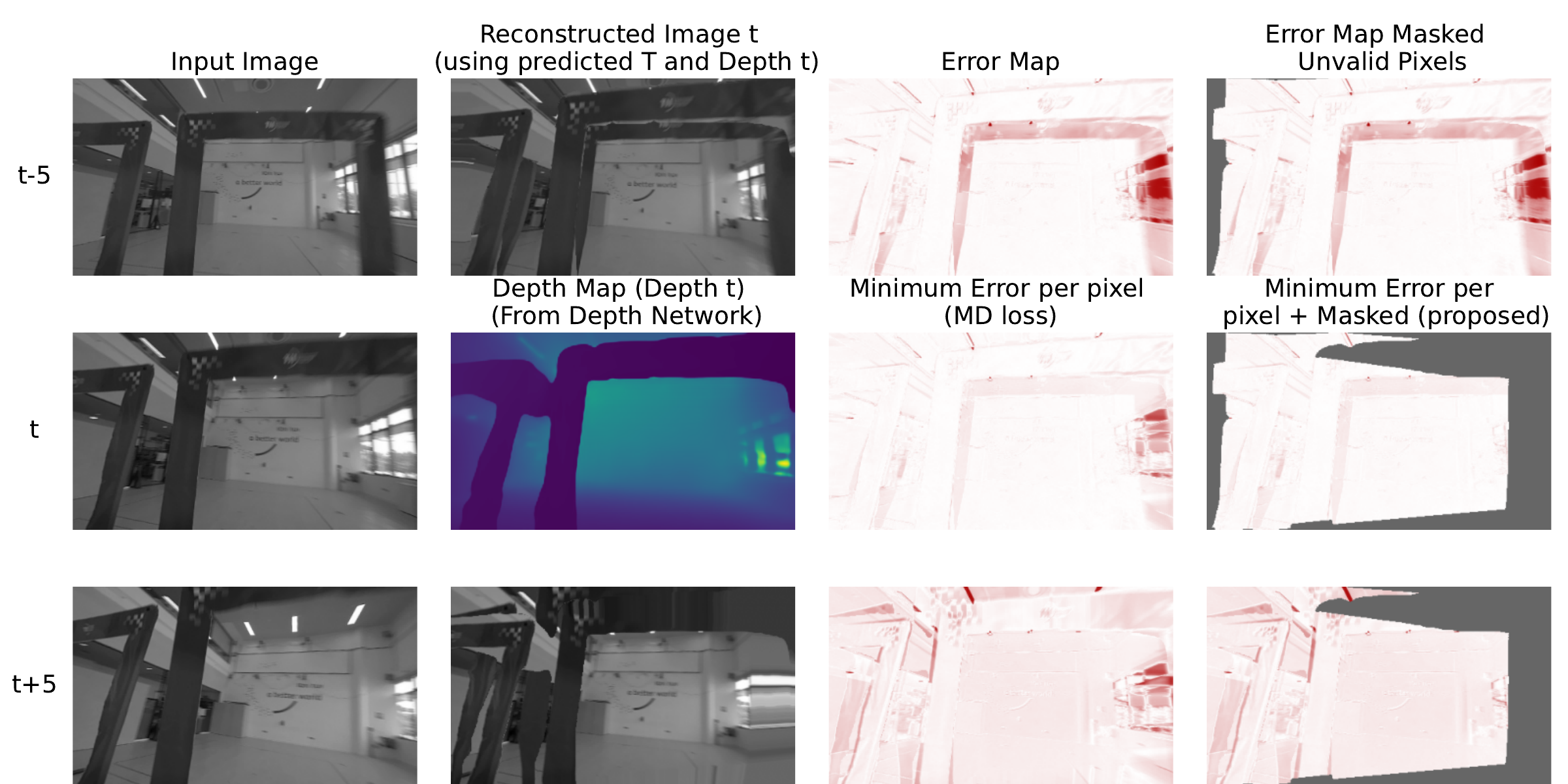}
    \caption{\textbf{Validity image reprojection:}
    Given three images at timestep $t-5$, $t$ and $t+5$, two reconstructions of image $I_t$ can be computed. By estimating a depth map at time t and two translation and rotation estimations, $T_{t-5\rightarrow t}$ and $T_{t+5 \rightarrow t}$. In the occluded parts, this reconstruction is undefined and creates artefacts.
    % Given an image at timestep $t$ called $I_t$, the corresponding estimated translation and rotatation $(T)_{t-5\rightarrow t}$ and estimated depth map $D_t$, a reconstruction can be computed of the image $I_{t}$. The same can be done for the reconstruction of the image $I_{t+5}$ using $(T,R)_{t\rightarrow t+5}$ which is called $R_{t+5}$. 
    % In the occluded parts, this reconstruction is undefined and creates artefacts.
    For instance, in the reconstructed image that uses image $t-5$ as input (top row), the pixels right and below of the gate can not be reconstructed as they were occluded in $I_{t-5}$. The reconstructed image using image $t+5$ as input (bottom row), cannot reconstruct the right edge of $I_t$ as they exited the frame. Both cases can theoretically not be correctly reconstructed, which leads to errors in the loss. This can be reduced by taking the minimum  \cite{monodepth2} as the errors are on the other side of edges (MD loss) is used. Pixels that move out of the image can be ignored using a valid pixel mask (light-gray, last column) as proposed by \cite{valid_pixels}. However, \cite{valid_pixels} only considered a single reconstruction (top/bottom row). We propose to use the minimum per valid pixel of both error maps that solves all edge cases, and we call it the \textbf{3F} 
    scheme.}
%    They proposed to use a valid pixel mask to take into account pixels moved out of the image, however they only considered two adjacent frames therefore in their loss function (

    %Reprojecting 2 adjacent frames can still result in an invalid reprojection error, combining it with the valid pixel mask from \cite{valid_pixels} solves the issue. In this figure, we skip 5 frames to improve the visibility of the problem.}
    \label{fig:occlusion_adjacentframes}
\end{figure*}

\begin{figure*}
    \centering
    \includegraphics[width=0.95\linewidth]{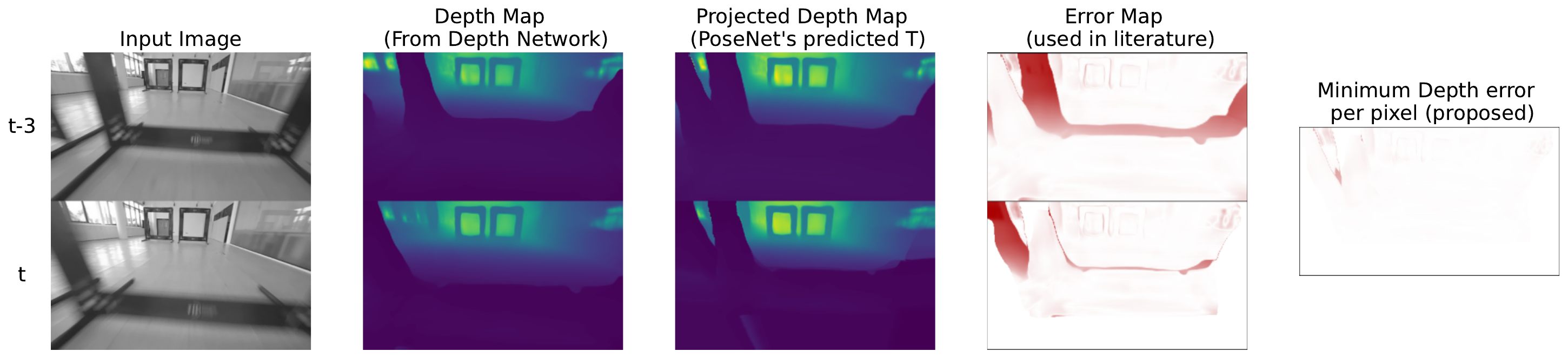}
    \caption{\textbf{Validity depth reprojection:} When approaching a gate, the reprojected depth map perceives similar artefacts as reprojected images due to occlusion. This causes an error in the depth consistency loss. In \cite{depth_consistency_loss} only one projection was considered and used the depth consistency loss based on the top Error Map. In our 2F method we reprojected $Dt$ as well (bottom row) using the inverse of the estimated relative pose (T), followed by taking the minimum of both reprojection errors (last column). Note that we also apply a valid pixel mask for depth consistency but do not show it here.}
    \label{fig:depth_consistency_error}
\end{figure*}
\section{Methodology}\label{section:method}

Our main methodology is as follows. First, we use self-supervised learning of a PoseNet and DepthNet, with the same network structure as Monodepth2 \cite{monodepth2} with a modified loss function. This monocular SSL scheme relies on the prediction of a target image from a source image. The main concept is that in a static environment, pixel displacements between the source and target image can be inferred from the depth and the relative camera pose between the two images (translation and rotation). Given the frame rate of the camera and the output of the SSL (Self-Supervised Learning) PoseNet, we get an unscaled velocity estimate in the camera frame. We then train a drone model, that estimates the specific forces that act on a drone (without external, wind disturbances) using inertial data and the rotor speeds. The unscaled velocities estimated by PoseNet serve as supervision. Given the attitude estimated from the IMU and the gravitational acceleration of $9.81\ m/s^2$, we can recover metric accelerations and hence find the scale parameter of PoseNet and DepthNet for a certain environment.

\subsection{PoseNetwork}
We use the Depth network that predicts a disparity map similar to Monodepth2 \cite{monodepth2} that is based on the U-net architecture \cite{u-net} with ResNet-18 \cite{resnet} as the encoder. The PoseNetwork is a separate network that has the same network structure as the depth network, without having skip connections.

Furthermore, for training our networks we use three different losses. Firstly, the appearance loss, which is a combination of the L1 loss and the Structural Similarity (SSIM) loss \cite{SSIM}. In Equation \ref{eq:appearance_loss} we show the appearance loss per pixel. 

% \begin{equation}\label{eq:appearance_loss}
%     \mathcal{L}_{\text {p}}(i,j)=(1-\alpha)\left|\mathbf{I}_{s \rightarrow t}(i,j)-\mathbf{I}(i,j)_t\right|+\alpha \mathcal{L}_{\mathrm{SSIM}}\left(\mathbf{I}_{s \rightarrow t}(i,j), \mathbf{I}_t\right(i.j))
% \end{equation}

\begin{equation}\label{eq:appearance_loss}
\begin{aligned}
    l_{\text{p}}(i,j) = & (1-\alpha) \left| \mathbf{I}_{s \rightarrow t}(i,j) - \mathbf{I}_t(i,j) \right| \\
    & + \alpha \mathcal{L}_{\mathrm{SSIM}} \left( \mathbf{I}_{s \rightarrow t}(i,j), \mathbf{I}_t(i,j) \right)
\end{aligned}
\end{equation}

% \begin{equation}\label{eq:appearance_loss}
%     \mathcal{L}_{\text {p}}(I_s,I_t)=(1-\alpha)\left|\mathbf{I}_{s \rightarrow t}-\mathbf{I}_t\right|+\alpha \mathcal{L}_{\mathrm{SSIM}}\left(\mathbf{I}_{s \rightarrow t}, \mathbf{I}_t\right)
% \end{equation}
% \left(\mathbf{I}_{s \rightarrow t}, \mathbf{I}_t\right)

% Similar to \cite{monodepth2}, \cite{dif_VIO_wagstaff} we use the disparity smooth loss from \cite{smooth_loss} as in Equation \ref{eq:smooth}. The disparity smooth loss encourage disparities to be locally smooth with an L1 penalty on the disparity gradients ∂d. As depth discontinuities often occur at image gradients,  the weight cost with an edge-aware term using the image gradients.

Next, we use a depth consistency loss that minimizes the difference of two consecutive predicted depth maps and the corresponding predicted relative pose. This loss tries to enforce scale consistency over a trajectory. In Equation \ref{eq:depthloss} we show the depth loss per pixel from  \cite{depth_consistency_loss}. It relies on the estimated transformation matrix between the frames from PoseNet:

% \begin{equation}\label{eq:depthloss}
%     \mathcal{L}_{\text {depth}}= \frac{1}{W \times H} \sum_{p=0}^{W \times H}  \frac{\left|D_{s \rightarrow t}(p)-D_{t}(p)\right|}{D_{s \rightarrow t}(p)+D_{t}(p)}
% \end{equation}

% \begin{equation}\label{eq:depthloss}
%     \mathcal{L}_{\text {D}}(D_s,D_t)= \frac{1}{N} \sum_{i,j}  \frac{\left|D_{s \rightarrow t}(i,j)-D_{t}(i,j)\right|}{D_{s \rightarrow t}(i,j)+D_{t}(i,j)}
% \end{equation}

\begin{equation}\label{eq:depthloss}
    l_{\text {D}}(i,j)= \frac{\left|D_{s \rightarrow t}(i,j)-D_{t}(i,j)\right|}{D_{s \rightarrow t}(i,j)+D_{t}(i,j)}
\end{equation}

Lastly,  we use the disparity smoothness loss from \cite{smooth_loss}, as defined in Equation \ref{eq:smooth}. The disparity smoothness loss encourages local smoothness in the predicted disparities by applying an L1 penalty on the disparity gradients, $\partial d$. Since depth discontinuities often coincide with image gradients, the loss is weighted using an edge-aware term that incorporates the image gradients, $\partial I$.

% \begin{equation}\label{eq:smooth}
%     \mathcal{L}_{\text {smooth}} = \left| \partial_x d_t^* \right| e^{-\left| \partial_x I_t \right|} + \left| \partial_y d_t^* \right| e^{-\left| \partial_y I_t \right|}
% \end{equation}

\begin{equation}\label{eq:smooth}
    \mathcal{L}_{\text {s}} = \frac{1}{N} \sum_{i.j} \left| \partial_x d_{i,j} \right| e^{-\left| \partial_x I_{i,j} \right|} + \left| \partial_y d_{i,j} \right| e^{-\left| \partial_y I_{i,j} \right|}
\end{equation}

% \begin{equation}
%     \mathcal{L}_{\text {total}} = \mathcal{L}_{\text {p}} + \lambda_1  \   \mathcal{L}_{\text {D}} + \lambda_2 \ \mathcal{L}_{\text {s}}
% \end{equation}

\subsection{Gate Occlusion}

In autonomous drone racing, the goal is to fly through a set of gates. Each time the drone flies through a gate, a large number of pixels move out of the image and a previously occluded area appears. This is problematic for SSL from reconstruction as the appearing pixels cannot be predicted from the previous image, but do lead to errors. Compared to car datasets like KITTI \cite{kitti}, occlusions and dis-occlusions (accretions) include more pixels in drone racing as the gates get closer to the camera and thereby affect the overall solution greatly. 
% CDW: zeker van? of gebeurt het DICHTER BIJ DE DRONE en is daarom MEER PIXELS: ik denk: KITTI heeft nog veel vaker OCCLUSIONS, maar dan kleine want ver-weg.
On the other hand, in autonomous drone racing datasets, there are currently no dynamic objects. Therefore, dynamic object masking networks are not necessary. 

In \cite{valid_pixels} a validity mask is used to remove pixels that moved out of the image from the loss function. However, (dis-)occlusion is not taken into account. In MonoDepth2 a solution is proposed that should solve both problems, by reprojecting both of the frames $I_{t-1}$ and $I_{t+1}$ to $I_{t}$ and take the minimum error per pixel location of both reprojections as in Equation \ref{eq:min_photo} \cite{monodepth2}. 

\begin{equation} \label{eq:min_photo}
    \mathcal{L}_{\text{p}} = \frac{1}{N} \sum_{i,j} \min \big( l_{\text{p1}}(i,j), l_{\text{p2}}(i,j) \big)
    % \min  \left( \mathcal{L}_{\text{p1}}(I_{t-1 \to t}, I_t), \mathcal{L}_{\text{p2}}\left(I_{t+1 \to t}, I_t\right) \right)
\end{equation}

The idea is that pixels that appear in $I_t$ due to disocclusion will still be visible in $I_{t+1}$. The loss map of this method is shown in Figure \ref{fig:occlusion_adjacentframes} as "MD loss". It can be observed that the error induced by the occluded area due to the left and top part of the gate in $I_{t-5}$ is corrected by the reprojection of $I_{t+5 \to t}$.

However, the right side of $I_t$ is occluded in $I_{t-5}$ and moves out of $I_{t+5}$ due to an aggressive yaw motion. As a result, an incorrect error is back-propagated when using the loss proposed in \cite{monodepth2}, which is commonly employed in self-supervised monocular depth and ego-motion networks, such as \cite{dif_VIO_wagstaff}. 

In \cite{valid_pixels} a valid pixel mask method is proposed to deal with pixels that move out of the image. However, in their approach, they only considered a single reprojection loss, $I_{t-5 \to t}$. The invalid pixels identified by this method are shown in gray in Figure \ref{fig:occlusion_adjacentframes} (last column). 
Therefore, we propose combining both methods to improve occlusion handling in self-supervised learning for ego-motion and mnonocular depth networks. The result is titled "proposed" and shown in Figure \ref{fig:occlusion_adjacentframes}, and is formulated in Equation \ref{eq:min_photo_masked}, where $V$ represents the valid pixel mask (set to $0$ if the pixel moves out of the image):

\begin{equation} \label{eq:min_photo_masked}
    \mathcal{L}_{\text{p}} = \frac{1}{N} \sum_{i,j} V_1(i,j) \ V_2(i,j)\ \min \big(  l_{\text{p1}}(i,j), l_{\text{p2}}(i,j) \big)
    % \min  \left( \mathcal{L}_{\text{p1}}(I_{t-1 \to t}, I_t), \mathcal{L}_{\text{p2}}\left(I_{t+1 \to t}, I_t\right) \right)
\end{equation}

Similar to image reconstruction, depth map reprojection also encounters occlusion-related issues. However, the depth consistency loss from \cite{depth_consistency_loss} (Eq. \ref{eq:depthloss}) does not account for occlusions, as it reprojects a single depth map from $t-1$ to the depth map at $t$. This leads to errors in occluded areas, such as when moving toward a gate, as illustrated in Figure \ref{fig:depth_consistency_error}. To address these occlusion issues, we apply a similar minimum depth reprojection and valid pixel masking approach. The formulation of this method is given in Equation \ref{eq:min_depth_valid}.

\begin{equation} \label{eq:min_depth_valid}
    \mathcal{L}_{\text{D}} = \frac{1}{N} \sum_{i,j} V_1(i,j) \ V_2(i,j)\ \min \big(  l_{\text{D1}}(i,j), l_{\text{D2}}(i,j) \big)
    % \min  \left( \mathcal{L}_{\text{p1}}(I_{t-1 \to t}, I_t), \mathcal{L}_{\text{p2}}\left(I_{t+1 \to t}, I_t\right) \right)
\end{equation}

The total loss function for the self-supervised PoseNet combines the photometric loss, depth consistency loss, and smoothness loss as in Equation \ref{eq:total_loss_vo}. Where $\lambda_1$ and $\lambda_2$ are weighting factors set to 0.15 and 0.001, respectively.

\begin{equation}\label{eq:total_loss_vo}
    \mathcal{L}_{\text {total}} = \mathcal{L}_{\text {p}} + \lambda_1  \   \mathcal{L}_{\text {D}} + \lambda_2 \ \mathcal{L}_{\text {s}}
\end{equation}

% valid pixel mask method proposed in \cite{valid_pixels}. This mask is

% such a pixel mask was also used, but only for the transformation from $I_{t-1}$ to $I_t$.

We investigate two alternative schemes for handling occlusions in self-supervised learning. 
In the first scheme, we use two adjacent source frames, $I_{t-1}$ and $I_{t+1}$ along with a target frame, $I_{t}$, and take the minimum reprojection error per pixel. This scheme is similar to MonoDepth2 \cite{monodepth2} (for the photometric loss). From hereon, we refer to this scheme as "3F" (three frames).

In the second scheme, we use only two frames, $I_{t-1}$ and $I_t$. These frames are reprojected onto each other using a single transformation (translation and rotation) and its inverse. Specifically, $I_{t-1}$ is first used as the source frame with $I_t$ as the target, and then the roles are reversed, with $I_t$ as the source and $I_{t-1}$ as the target. We refer to this scheme as "2F" (two frames). We hypothesize that applying the estimated transformation twice provides a stronger constraint on the solution, particularly when accounting for occlusions and disappearing pixels.

\subsection{Learned Drone Dynamics}
\begin{figure*}
    \centering
    \includegraphics[width=0.98\linewidth]{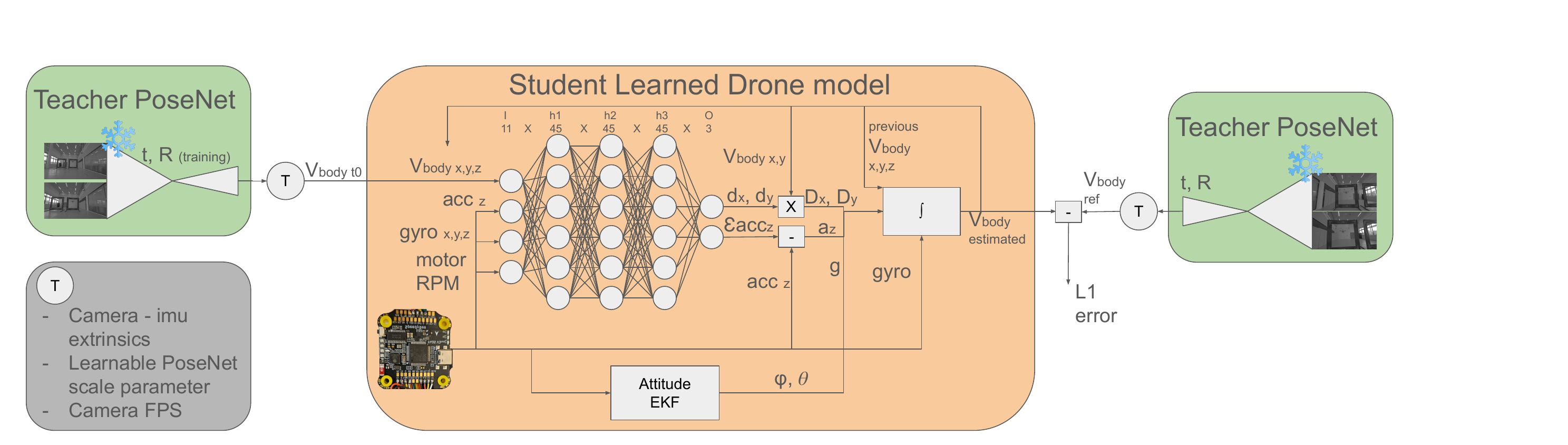}
    \caption{Single learnable parameter (in T) is used to transform the scaleless PoseNet velocity to a scaled body velocity. This is then used together with the IMU and motor RPMs to estimate a residual on the accelerometer-z and drag. By integrating the specific forces and the gravitational acceleration of $9.81 m/s^2$ the drone model is forced to recover the scale correctly}
    \label{fig:drone_model_overview}
\end{figure*}

Once the self-supervised PoseNet is trained, we can obtain an estimate of the drone's velocity by multiplying the predicted translation by the camera's frame rate (FPS). However, PoseNet is inherently scale-ambiguous, as scale cannot be directly observed from monocular vision. Despite this limitation, the depth consistency loss in PoseNet encourages a consistent scale across the predictions of a trajectory. As a result, PoseNet's velocity estimates should be proportional to the true scale by a single unknown factor. This allows us to still use PoseNet as a teacher to train the drone model, by simultaneously learning a scaling parameter to correct the scale. Furthermore, the camera-IMU extrinsics are required to transform the velocity from the camera frame to the body frame, assuming the IMU is located in the body frame. With this transformation, the body-frame velocities from PoseNet can be computed as shown in Equation \ref{eq:body_vel}.

\begin{equation}\label{eq:body_vel}
    V_{b_{pnet}} = s \cdot R_{c \rightarrow b} \bar{V}_{c}
\end{equation}

% \begin{equation}\label{eq:body_vel}
%     V_{b_{pnet}} = s \cdot R_{c \rightarrow b}(\Delta\phi, \Delta\theta, \Delta\Psi) \bar{V}_{c}
% \end{equation}

\noindent in which $s$ is a single learnable parameter to correct the scale of PoseNet across the entire trajectory. Thus, for $n$ training sequences, there are $n$ separate learnable scaling parameters.% Additionally, it is possible to estimate or fine-tune the camera extrinsics by introducing three more learnable parameters, $\Delta\phi, \Delta\theta, \Delta\Psi$, which adjust the transformation between the camera and body frame. However, we observed that allowing these parameters to be freely optimized throughout training can lead to suboptimal results, as they may be incorrectly adjusted due to imperfections in PoseNet’s predictions. To mitigate this, one could freeze these parameters early in the training process or perform a manual IMU-Camera calibration, like Kalibr \cite{kalibr}. Furthermore, both the input selection and the neural network architecture must be carefully designed to prevent overfitting to incorrect velocity estimates.

The primary forces acting on the drone are thrust, drag, and gravity. For the drag forces we assume that external disturbances (from wind for example) are negligible, because we are considering indoor environments only. Furthermore, we assume that thrust is aligned exactly with the body z-axis (upward direction), as we consider a quadrotor in this work. Under these assumptions, the sum of forces can be expressed as shown in Equation \ref{eq:sumforces}.

\begin{equation}\label{eq:sumforces}
    \sum \mathbf{F} = T_z + \mathbf{D} + m \mathbf{g}
\end{equation}

\noindent where $\mathbf{g}$ is the gravity vector, $T_z$ the thrust and $\mathbf{D}$ drag.
The gravity vector is computed by estimating the drone's attitude (roll and pitch) using an Extended Kalman Filter (EKF) on IMU data (accelerometer and gyroscope). The EKF implementation is similar to the one used in BetaFlight flight controllers \cite{betaflight}, where the gyroscope measurements are integrated, and an update step is performed when the accelerometer readings are in the range $[0.95, 1.05]\ g$.
%Alternatively, for short flights, one can estimate the gyro bias before the flight and integrate the bias-corrected gyro data, by using the the estimated (constant) gyro biases. 

The thrust generated by a quadcopter is a function of the relative airspeed of the rotor blades and their angle of attack (AoA), both of which depend on the rotor RPM ($\omega$) and the body velocities ($V_b$). Additionally, the relative airspeed of each rotor is slightly influenced by the yaw rate. For example, when the quadrotor rotates to the left, the advancing blade on the right side of the drone experiences a slightly higher airspeed than the retracting blade. The opposite occurs on the left side. Similarly, the AoA is affected by the pitch and roll rates, further influencing the thrust dynamics.

% The thrust of a quadcopter is a function of the relative airspeed to the blade and angle of attack (AoA), which both are functions of rotor RPM ($\omega$), and body velocities ($V_b$). Furthermore, the relative airspeed of a rotor also slightly depends on the yaw rate, for example when the quadrotor rotates to the left, the advancing blade of the right side of the drone has a slightly higher airspeed than the retracting blade. For the left side, it's the opposite. Similarly, the AoA is affected by the pitch- and roll rate.
%e.g. if the drone has a downward pitch rate,  the advancing blades have an increased AoA and the retreating blades a decreased angle of AoA. Not sure if this is correct, I need to think over this a bit better

The drag consists of two parts, body drag and rotor drag. The body drag has a quadratic relationship with the body velocity and is typically small and negligible compared to rotor drag at low flight speeds (below $5 m/s$) \cite{silva2018quadrotor}. Rotor drag, on the other hand, is more complex and consists of induced drag. This occurs when a drone is moving with a certain velocity. One blade of a rotor experiences an increased incoming airflow, while the other experiences a reduced incoming airflow. Since drag is proportional to velocity squared, the induced drag has a linear relationship to body velocity. Additionally, this effect not only increases drag but also alters the generated lift, causing the rotor to flap and further tilt the lift vector. Moreover, during aggressive turns, the drone can fly into its own wake, leading to turbulent airflow. This turbulence significantly impacts the lift and drag forces on the quadrotor and is difficult to model accurately.

Since it is hard to explicitly measure and model the effects described above, we train a neural network to exploit the information available onboard the drone. The inputs to the network include (its previously) estimated body velocity, the accelerometer z-axis, gyroscope x, y, z, and the four motor RPM feedback signals. With these inputs, the neural network is designed to predict the sum of the specific forces, excluding gravity. The gravity component is precomputed for each time sample using the IMU EKF attitude estimation.

After removing gravity, the specific forces in the body x and body y directions are solely due to drag. Since drag is directly dependent on body velocity, we output two drag terms: $d_x$ and $d_y$, which are multiplied by the respective body velocities $V_{bx}$ and $V_{by}$ to represent the specific force in x and y. By having the neural network estimate the drag terms instead of directly predicting the specific drag force, the learning process becomes more stable. When velocity decreases, the estimated drag naturally decreases as well, preventing oscillations and avoiding the risk of predictions diverging over long open-loop prediction periods. The output of $d_x$ and $d_y$ is limited to [$0.0,\ 2.0$].

However, this approach cannot be applied to the z-axis, as the specific force in this direction is influenced by both drag and thrust. The accelerometer measures the combined effect of these forces, so instead of predicting thrust and drag separately, the neural network is trained to predict the residual error of the accelerometer, $\varepsilon_{acc_z}$ in the z-axis. Since the accelerometer in the z-axis is an input to the network itself, the network has the potential to implicitly learn the underlying relationship to thrust and drag in the z-axis. The output range of $\varepsilon_{acc_z}$ is constrained to [$-5.0,\ 5.0$]. 

An overview of the structure can be seen in Figure \ref{fig:drone_model_overview}. We use a feedforward neural network with three fully connected hidden layers of size $[45,45,45]$. The hidden layers use a $tanh$ activation function, and the output layer of size 3 uses the sigmoid activation function, properly rescaled to the relevant intervals. The body velocity is estimated with Equation \ref{eq:V_est}:

% By doing this we ensure that we are not destroying the scale measurement of gravity. Using the estimated specific forces and the measured rotation from the gyro we can estimate the new body velocity as in Equation \ref{eq:integrate_vel}. Note that for each timestamp we rotate the velocity with the opposite rotation of the measured gyro to correct for rotation to the body velocity.

\begin{equation}\label{eq:V_est}
    Vb_{t} = \sum_{i=0}^t  R_{\omega \Delta t_i}^T \left(  Vb_i + 
    \begin{bmatrix}
    -d_x Vb_{xi} + g_x \\
    -d_yVb_{yi} + g_y \\
    a_z - \varepsilon_{a_{z}} + g_z
    \end{bmatrix} \Delta t_i
    \right)
\end{equation}

in which $R_{\omega \Delta t_i}^T$ is the inverse of the estimated body rotation from the gyroscope during $\Delta t_i$, $g$ is the estimated gravity vector of the attitude EKF, $a_z$ is the measured acceleration. During training, $Vb_0$ is initialized with PoseNet's estimate to allow data shuffling. For evaluation, it is initialized as zero at the start of a trajectory and relies on open-loop predictions.

% \begin{equation}\label{eq:integrate_vel}
%     V_{b_{t+1}} = R^T(\omega_x, \omega_y, \omega_z, dt) \left ( V_{b_t} + acc_t \cdot dt  \right)  
% \end{equation}

% \begin{equation}
%     \mathcal{L} = | Vb_{t+skip} - \sum_i^{skip}  R^T(\omega, dt) \left(  Vb_i + 
%     \begin{bmatrix}
%     d_x Vb_{xi} \\
%     d_yVb_{yi} \\
%     a_z - \varepsilon_{a_{z}}
%     \end{bmatrix} dt_i
%     \right)
% \end{equation}

% Alternatively, one can estimate the attitude by integrating the bias-corrected gyro data, by using the the estimated (constant) gyro biases from \cite{eq:gyroloss}, for short flights. 

\section{Results}\label{section:results}

First, we present the results of our improved occlusion handling for the self-supervised PoseNet loss. Next, we train a neural drone model that learns from PoseNet’s output and demonstrate that its relative velocity error decreases compared to its teacher. Finally, we integrate this neural drone model into ROVIO \cite{ROVIO}, and show that it enhances state estimation in aggressive racing flights.

\subsection{Dataset}

We train and test our method on the TII Drone Racing Dataset \cite{tii}, which was recorded in an indoor environment where an autonomous drone flies at speeds of up to $22\ m/s$. The dataset includes monocular camera recordings at $120\ Hz$, along with IMU data and motor RPM feedback recorded at $500\ Hz$. Currently, this is the only publicly available dataset that enables us to demonstrate how a neural drone model can improve state estimation, as our method relies on both high-speed flight and motor RPM feedback.

The dataset includes 12 piloted flights, with $6$ elliptical trajectories ($01p-06p$) and $6$ lemniscate trajectories ($07p-12p$), each lasting $\sim 100$ seconds. Additionally, there are $18$ autonomous flights, consisting of $6$ elliptical ($01a-06a$), 6 lemniscate ($07a-12a$), and $6$ $3$D racing trajectories ($13a-18a$), with durations ranging from $10$ to $20$ seconds per flight.

We preprocess the data by undistorting the images and converting them to grayscale for all tests, including for ROVIO. To speed up training, PoseNet uses resized images of $448 \times 256$ pixels. We split the dataset into training, validation, and test sets as follows. Training set: The first $4$ flights of each trajectory type ($1-4$ elliptical, $7-10$ lemniscate, $13-16$ 3D). Validation set: Flights $6$, $12$, and $18$. Test set: Flights $5$, $11$, and $17$.

% \begin{itemize}
%     \item Training set: The first $4$ flights of each trajectory type ($1-4$ elliptical, $7-10$ lemniscate, $13-16$ 3D).
%     \item Validation set: Flights $6$, $12$, and $18$.
%     \item Test set: Flights $5$, $11$, and $17$.
% \end{itemize}

\subsection{Self Supervised Visual Ego-motion Estimation}

We trained an iterative PoseNet similar to \cite{dif_VIO_wagstaff}, but we limited the number of iterations to $3$ instead of $5$ to reduce training time. We used a learning rate of $5 \cdot 10e^{-5}$ with a decay of $10\%$ every $8$ epochs and for the Adam optimizer we set $\beta_1=0.9$ and $\beta_2=0.999$. The depth consistency loss and smooth loss have a weight of $0.15$ and $1e-3$, respectively. A batch size of $60$ for the $2$-frame loss, while a batch size of $48$ was used for the $3$-frames loss. This difference is due to memory constraints, as using only two images requires less memory, allowing for a larger batch size. For data augmentation, we augment the brightness and contrast, and we randomly flip the image horizontally. We also randomly skip $0-3$ frames. For the 3D racing track trajectories we only skip $0-2$, because those tracks are more aggressive flight than the others.

We compare the iterative PoseNet trained with the 2F and 3F method. As benchmark we train an additional network that follows the approach of \cite{dif_VIO_wagstaff} and use the Monodepth2 \cite{monodepth2} photometric loss, depth consistency loss from \cite{depth_consistency_loss} and the disparity smooth loss \cite{smooth_loss}. For comparison, we integrate the pose outputs as a straightforward form of visual odometry, leading to trajectories that can be aligned and compared with the ground-truth trajectories. We observed that the trajectory estimations of all PoseNets heavily changed per epoch due to incorrect attitude estimation, especially for the piloted flights that have a long duration. Therefore, we stop training based on the RMS of the absolute position error of the validation trajectories. For each network we evaluate the epoch that had the lowest average absolute position root-mean square error (RMSE) of all validation trajectories.%, see Figure \ref{fig:val_rmse}. 
% The lowest average RMSE is marked with a star and is $2.38\ m$ for the benchmark network, $2.06\ m$ for the $3$ frames network and $1.88\ m$ for the PoseNet trained with 2 frames.

% Furthermore, we compare the results of the self-supervised PoseNets with the traditional method ROVIO \cite{ROVIO}. We test ROVIO without tuned and with tuned parameters using the ground truth (GT). In the TII dataset the IMU and images are not hardware synchronized, we empirically estimated the average delay to be $12\ ms$ which was required to run ROVIO successfully.

% The result of ROVIO and the self-supervised iterative PoseNets can be seen in Table \ref{tab:RMSE_VO}. ROVIO-0 are the original settings from \cite{ROVIO} and we show the results of the tuned setting with ROVIO*. It can be seen that ROVIO required tuning to work properly, but in return, it resulted in the most accurate trajectory estimation for the elliptical and lemniscate trajectories. However, for the more aggressive race track the self-supervised VO was more accurate and more robust than ROVIO. For flight 13 ROVIO* the VIO diverged when using an IMU delay of $12\ ms$, but when using $8\ ms$ the filter did not diverge and got an RMSE of $4.84\ m$. This shows that ROVIO is sensitive to IMU delays on aggressive flights. Furthermore, we noticed that the trajectory estimate of ROVIO was drifting (while it was able to track the features correctly) when it was flying slowly backwards to start position of the race tracks 13-18A. We took this part of trajectory out for the RMSE in Table \ref{tab:RMSE_VO}. 

The absolute position RMSE is presented in Table~\ref{tab:RMSE_VO}. Since PoseNet predicts scale-ambiguous transformations from monocular images, we apply a 7-DoF (SIM3) alignment following \cite{umeyama} to resolve scale inconsistencies. We observe that the 2F approach achieves a lower RMSE than the 3F approach for nearly all trajectories, with an average reduction of $25\%$. Additionally, training is faster for the 2F method, as the transformations for the two reprojections are simply the inverse of each other. In contrast, the 3F method requires computing transformations from $t-1$ to $t$ and from $t+1$ to $t$ for each training sample, increasing computational complexity. However, this comparison is conducted in a static environment. In dynamic environments, the 3F approach may be advantageous, as the additional frame could help in mitigating the influence of moving objects on the loss. Furthermore, we observe that the 2F method achieves an average RMSE reduction of $15\%$ compared to the benchmark network. Surprisingly, our 3F method, which incorporates the improved depth and photometric loss, results in a higher absolute position RMSE than the benchmark on average. 

\begin{table}[h]
\caption{Absolute position RMSE in meters of self-supervised PoseNets after 7dof (SIM3) alignment. }\label{tab:RMSE_VO}
\begin{center}
\begin{tabular}{lccc}
\hline
         & \multicolumn{1}{l}{\begin{tabular}[c]{@{}l@{}}Bench-\\ mark\end{tabular}} & \multicolumn{1}{l}{\begin{tabular}[c]{@{}l@{}}3F\\ (ours)\end{tabular}} & \multicolumn{1}{l}{\begin{tabular}[c]{@{}l@{}}2F\\ (ours)\end{tabular}} \\ \hline
01-04P   & 3.64                                                                      & 4.18                                                                    & \textbf{1.92}                                                           \\
05P (T)  & 1.94                                                                      & 3.72                                                                    & \textbf{1.87}                                                           \\
06P (V)  & 2.35                                                                      & 2.20                                                                    & \textbf{1.33}                                                           \\
07-10P   & \textbf{3.81}                                                             & 4.25                                                                    & 3.96                                                                    \\
11P (T)  & \textbf{4.79}                                                             & 5.45                                                                    & 4.86                                                                    \\
12P (V)  & 4.91                                                                      & \textbf{4.00}                                                           & 4.07                                                                    \\
01-04A   & 0.85                                                                      & \textbf{0.81}                                                           & 0.89                                                                    \\
05A (T)  & \textbf{0.90}                                                             & 1.03                                                                    & 1.08                                                                    \\
06A (V)  & 0.98                                                                      & 0.87                                                                    & \textbf{0.65}                                                           \\
07-10A   & \textbf{0.75}                                                             & 0.91                                                                    & 1.11                                                                    \\
11A (T)  & \textbf{0.90}                                                             & 1.41                                                                    & 1.11                                                                    \\
12A (V)  & 1.41                                                                      & \textbf{1.18}                                                           & 1.25                                                                    \\
13-16A   & 2.71                                                                      & \textbf{2.11}                                                           & 2.14                                                                    \\
17A (T)  & 3.09                                                                      & 2.77                                                                    & \textbf{2.35}                                                           \\
18A (V)  & 2.25                                                                      & \textbf{2.04}                                                           & 2.11                                                                    \\ \hline
Avg      & \multicolumn{1}{l}{2.35}                                                  & \multicolumn{1}{l}{2.51}                                                & \multicolumn{1}{l}{\textbf{2.00}}                                       \\
Std dev. & \multicolumn{1}{l}{1.40}                                                  & \multicolumn{1}{l}{1.54}                                                & \multicolumn{1}{l}{\textbf{1.19}}                                       \\ \hline
\end{tabular}
\end{center}
\end{table}

\begin{figure}[htb]
    \centering
    \includegraphics[width=0.9\linewidth]{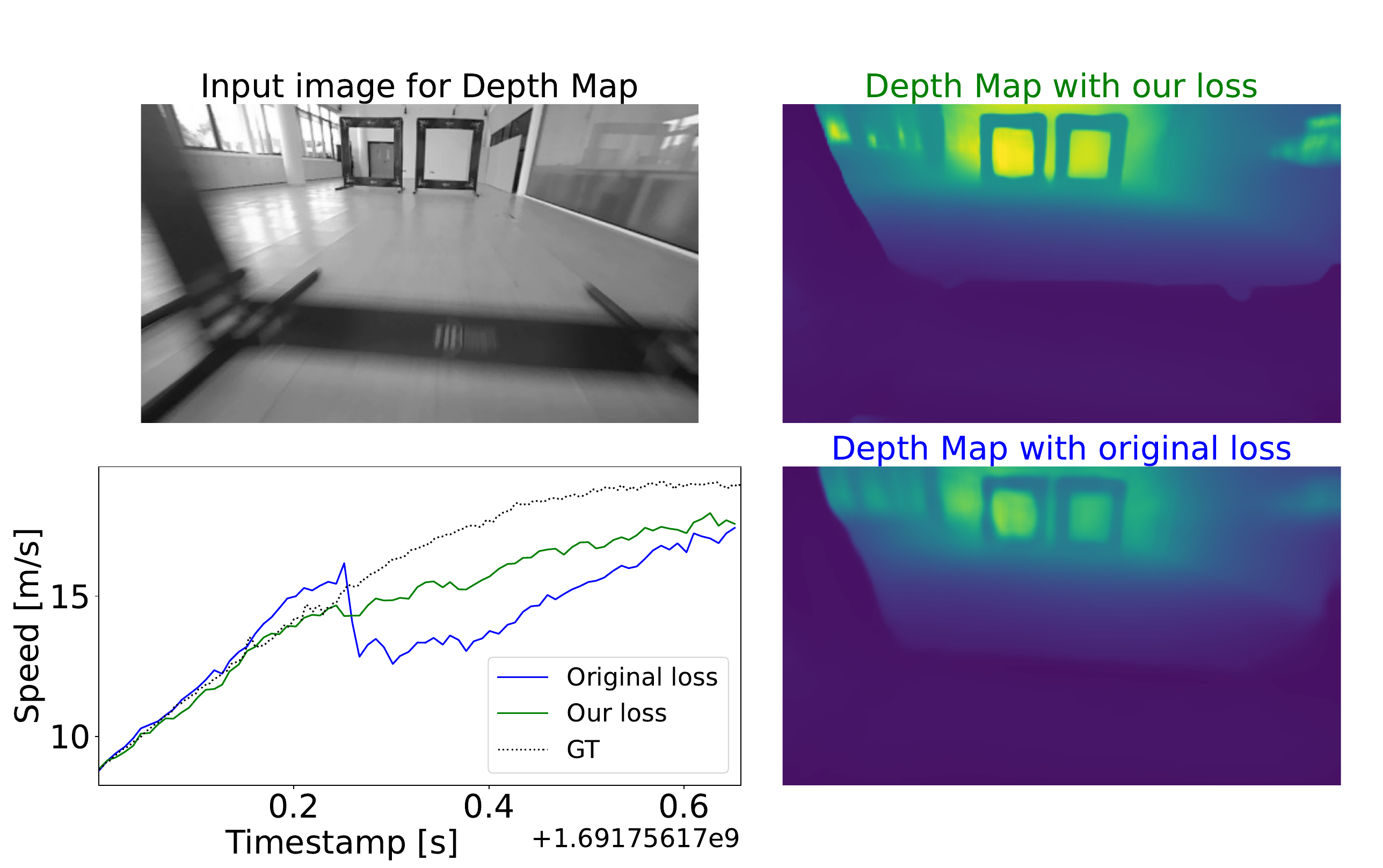}
    \caption{Our loss function improves both depth and velocity estimation when there are large (dis-)occlusions. Top left: input image when passing through a gate. Bottom left: speed, estimated with motion tracking (GT, grey), and with a PoseNet trained with our loss (green) and the original loss (blue). Top right: Depth map from PoseNet with our loss, from close by (dark blue) to far (bright yellow). Bottom right: Depth map from PoseNet trained with the original loss.}
    \label{fig:scale_consistency_2f}
\end{figure}

To demonstrate that our loss improves self-supervised learning (SSL), we provide a qualitative comparison of the results with the original loss and our proposed loss when flying through a gate, as shown in Figure \ref{fig:scale_consistency_2f}. We observe that the depth map obtained using our loss is sharper. This is due to the better occlusion handling, reducing the impact of erroneous backpropagation from occluded areas behind the gate. In contrast, the original method struggles with these occlusions, leading to an underestimated background depth, which in turn causes an underestimated translation (velocity) estimate. By improving translation consistency, our approach results in a more accurate PoseNet teacher, which in turn enables better training of our student neural drone model.

% The absolute position RMSE can be seen in Table \ref{tab:RMSE_VO}. Note that we apply a 7dof (SIM3) alignment \cite{umeyama}, because the output of PoseNet is scale-less since it is not observable from monocular images only. Comparing the 3F and 2F method, we see that the 2F method has a lower RMSE than the 3F method for almost all trajectories with an average of $25\%$ smaller RMSE. Furthermore, a single training epoch is faster for the 2F method because the transformation from source to target and target to source is its inverse. Whereas, from the 3F method the transformation from $t-1$ to $t$ and $t+1$ to $t$ needs to be computed for a single training item.

% Furthermore, we see that the 2 frames (2F) method has a lower RMSE than the 3 frames (3F) and the benchmark network and on average has a $15\%$ lower RMSE than the benchmark. Unexpectedly, the 3F occlusion aware has a higher trajectory error than the benchmark.  In Figure \ref{fig:scale_consistency_2f} we show that with 2F has an improved scale consistency when approaching the gate comparing with the benchmark (labeled as original loss). The reason is that in the depth consistency loss the occlusion was not taken into account, as we showed in Figure \ref{fig:depth_consistency_error}.

\subsection{Self Supervised Drone Model}
We precompute PoseNet estimates and divide by the time interval between the two input images to obtain scale-less velocity estimates. Since some frames are missing in the TII drone racing dataset, we use the PoseNet output to detect these gaps and correct the corresponding time intervals. Before using the velocity estimates, we apply a third-order Butterworth filter (forward and backward) to smooth the data. Additionally, we observe that the drone model trains better when using longer training samples, as this reduces the influence of PoseNet estimation errors. We use a varying training sample between $0.25 - 5 \ s$. The training, validation, and test sets remain consistent with our previous experiments.

% % \ref{fig:5a}, 
% In Figure \ref{fig:17a} we show the velocity estimates of the learned drone model and PoseNet on the 3D race trajectory (from the test set) using the estimated scale parameter we got from learning the drone model. The scale of PoseNet is recovered by taking the average of the four similar trajectories that are in the training set (13a-16a in this case). Due to depth consistency loss all trajectories in the same environment should have a similar scaling parameter. The average scale parameter estimated (from PoseNet to world) is 7.08 with standard deviation of 0.24 (over the 20 trajectories in the training set). Even though, the velocity estimates of the learned drone model are quite accurate, the learned drone model drifts over time as can be seen in Table \ref{tab:RMSE_VO}. We can see that especially for the longer trajectories the position error drifts, which is as expected since the model only relies on IMU, motor RPMS and body velocity. Therefore, it is stable and accurate in velocity estimation but drifts in position. However, for the short autonomous flights, like the elliptical flights it was almost as accurate as PoseNet itself.

\begin{figure}[htb]
    \centering
    \includegraphics[width=0.99\linewidth]{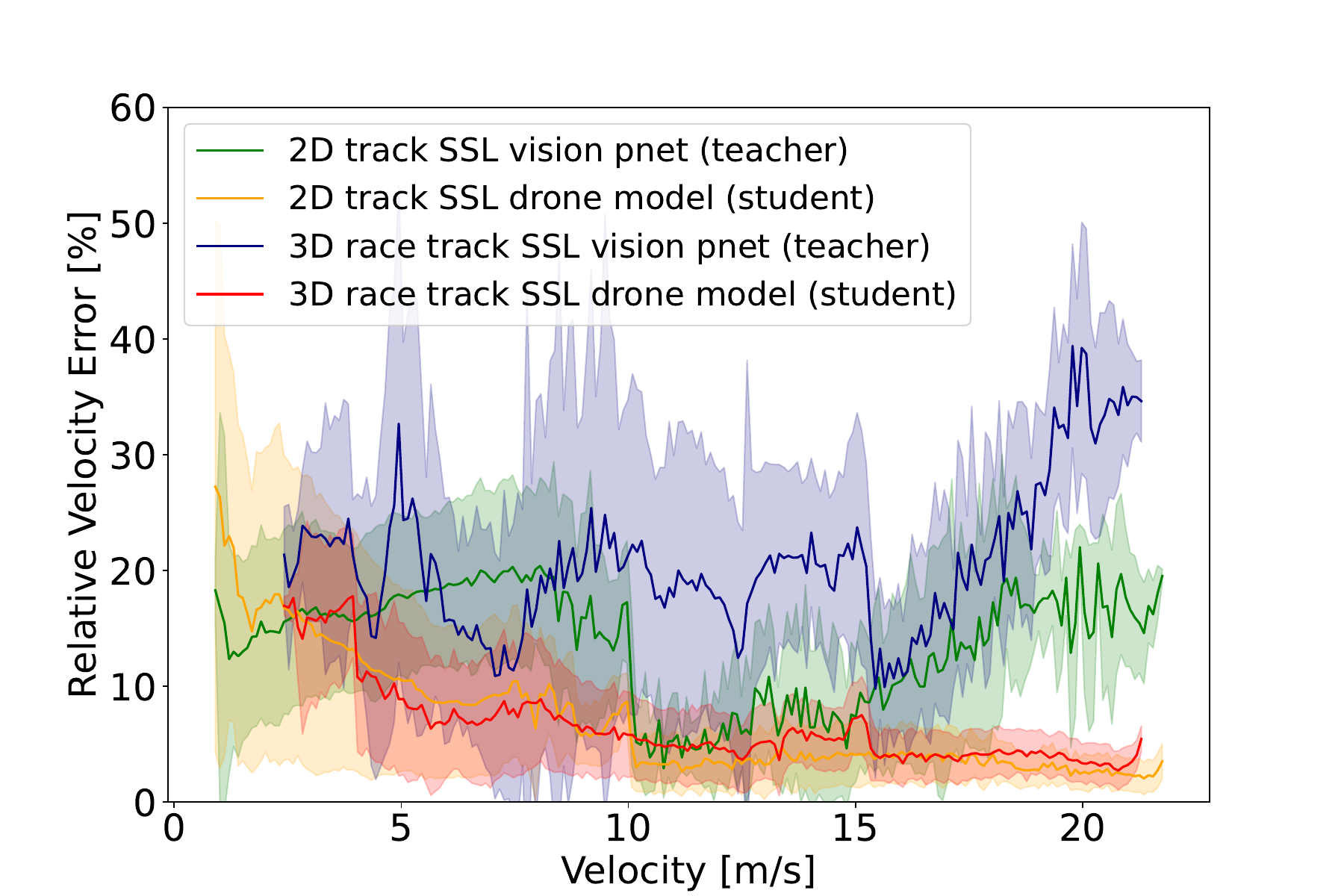}
    \caption{Relative Velocity Error vs. Flight Speed for Self-Supervised Vision PoseNet (Teacher) and the standalone open-loop predictions of the Self-Supervised Neural Drone Model (Student) on the TII Drone Racing Dataset. The Solid Line Represents the Mean, and the Shaded Area Indicates $\pm1$ Standard Deviation. It shows that at high speeds the drone model can predict the speed much more accurately than vision can measure it.}
    \label{fig:velocity_error}
\end{figure}

In Figure \ref{fig:velocity_error}, we plot the relative velocity error against speed. We observe that the learned drone model (student) has a decreasing relative velocity error at higher speeds, whereas the error for PoseNet (the teacher) increases. This indicates that the learned drone model outperforms its teacher, PoseNet, in velocity estimation. However, it is important to note that PoseNet estimates translation and rotation, while here we evaluate velocity by simply dividing the translation by $\Delta t$ between two frames. Furthermore, we observe that PoseNet performs worse on the 3D race track compared to the 2D track. This degradation occurs because, in the 3D race track, the drone undergoes more rotations, leading to increased motion blur, whereas motion blur is less pronounced in the 2D track. Since motion blur does not affect the learned drone model, we observe less degradation in its performance.

Unlike PoseNet, which directly estimates translation, the neural drone model predicts acceleration. Velocity estimates for the drone model are obtained through open-loop integration, where each trajectory starts with an initial velocity of $V_x, V_y, V_z = 0$, and the velocity is computed by integrating the neural drone model’s acceleration predictions along with gyroscope data. Because the neural drone model provides accurate motion predictions, it can be integrated into filtering techniques to enhance state estimation.

We integrated the neural drone model with the state-of-the-art filtering-based visual-inertial odometry (VIO) method, ROVIO. Specifically, we estimate the drone's acceleration using a weighted combination of the neural drone model ($30\%$) and the IMU accelerometer ($70\%$). We then compare this hybrid approach to the standard ROVIO implementation, which relies solely on the IMU for acceleration estimation. %This fusion strategy is similar to the winning algorithm of the AIIR drone racing competition in 2019 \cite{mavlab-winning}, which used a parameter-estimated linear drag model for improved state estimation.

\begin{figure*}
    \centering
    \includegraphics[width=0.99\linewidth]{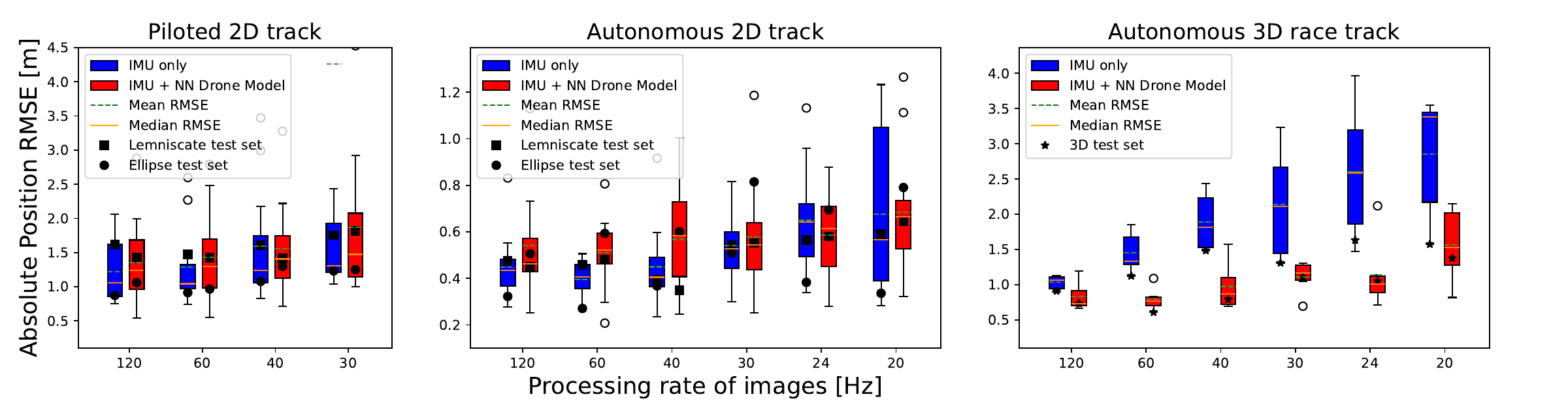}
    \caption{Absolute RMSE (after 6DOF alginment) vs. Skipping Frames Using Traditional ROVIO and a Neural Drone Model Hybrid Approach. The Hybrid Method Performs Better in Aggressive Flights (Autonomous 3D race track).}
    \label{fig:boxplot}
\end{figure*}

Figure \ref{fig:boxplot} shows the absolute RMSE. The x-axis represents the processing rate, where $120\ Hz$ indicates no skipped frames, $60\ Hz$ means every other frame is skipped, $40\ Hz$ skips two frames and processes one, and so on. The piloted 2D boxplot includes trajectories 01p–12p (12 trajectories), autonomous 2D includes 01a–12a, and autonomous 3D includes 13–18.  Additionally, the RMSE of the trajectory in the test set is highlighted with a special marker in the boxplot.

We observe that the primary performance difference occurs in the 3D race track, where the flight is more aggressive and visual conditions are more challenging. In this scenario, incorporating $30\%$ neural drone model predictions significantly improves state estimation, highlighting its effectiveness in demanding environments. Moreover, the performance is more pronounced for lower processing rates, when the odometry relies more on the inertial and model information. 

Furthermore, for the piloted flights, we only present results up to 30 Hz because, beyond this rate, some piloted trajectories began to diverge. This issue arises because the dataset captures the drone taking off almost immediately, which can prevent ROVIO from properly initializing, especially when frames are also skipped. In these cases, the neural drone model helps stabilize VIO, as an overestimated velocity leads to an overestimated drag, counteracting potential drift. As a result, at $30\ Hz$, the mean RMSE for the $30\%$ neural drone model + IMU combination is lower than using only the IMU.

\section{Conclusion}\label{section:discussion}
We introduced a novel self-supervised learning approach for ego-motion estimation, based purely on information available onboard a drone. We improved occlusion handling for self-supervised PoseNets, leading to more accurate depth estimation and, consequently, more scale-consistent translation predictions. Building on this, we introduced the first self-supervised neural drone model, trained as a student network with PoseNet as the teacher. Our results show that the neural drone model outperforms its teacher, PoseNet, in estimating body velocities, particularly at high speeds. Unlike PoseNet, which suffers from increasing errors at high rotational velocities due to motion blur, the neural drone model remains robust and is barely affected by the aggressive rotations which are common in drone racing scenarios. Finally, we integrated the neural drone model into the state-of-the-art filter-based VIO method, ROVIO, and demonstrated that it improves state estimation in challenging 3D race trajectories.

% Traditionally, motion captures systems or other external ground truth sources were required for learning drone models, making the process constrained to lab environments. In contrast, our method only requires a camera and flight controller data, making it far more scalable and accessible, since most drones are already equipped with a camera.

% We have trained a self-supervised drone model using onboard flight controller data and a monocular video. In agile flights, the velocity estimates from the drone model are more consistent than the teacher, PoseNet. However, the drone model does drift as can be seen in Table \ref{tab:RMSE_VO}. Therefore, the next step would be to fuse both networks to get the best of both of them. The fusion of both networks could be a stepping stone to deploy agile autonomous drones into the real world. Furthermore, the learned drone model is scale aware and fusing both networks may be a solution to the scale issue of PoseNet in unseen environments. Therefore, we will try to do this in our future work. 

% We also have seen that projecting two adjacent frames to one target frame was less accurate than projecting two frames to each other. Considering that it has been shown that it works well with the photometric error in \cite{monodepth2}. We suspect that this has to do with smoothness of the depth maps. Because, taking the minimum error of two adjacent depth maps (without a transformation) results in a small error already, leading to learning less.

\addtolength{\textheight}{-12cm}   % This command serves to balance the column lengths

%%%%%%%%%%%%%%%%%%%%%%%%%%%%%%%%%%%%%%%%%%%%%%%%%%%%%%%%%%%%%%%%%%%%%%%%%%%%%%%%

%%%%%%%%%%%%%%%%%%%%%%%%%%%%%%%%%%%%%%%%%%%%%%%%%%%%%%%%%%%%%%%%%%%%%%%%%%%%%%%%



% \section*{ACKNOWLEDGMENT}

% The preferred spelling of the word ÒacknowledgmentÓ in America is without an ÒeÓ after the ÒgÓ. Avoid the stilted expression, ÒOne of us (R. B. G.) thanks . . .Ó  Instead, try ÒR. B. G. thanksÓ. Put sponsor acknowledgments in the unnumbered footnote on the first page.

%%%%%%%%%%%%%%%%%%%%%%%%%%%%%%%%%%%%%%%%%%%%%%%%%%%%%%%%%%%%%%%%%%%%%%%%%%%%%%%%

\bibliographystyle{IEEEtran}
\bibliography{root}

% \begin{thebibliography}{99}

% \end{thebibliography}

\end{document}